\documentclass[twoside,11pt]{article}

\usepackage{pgm}
\usepackage{amsmath}
\usepackage{cancel}
\usepackage{xfrac}
\usepackage{multirow}

\ShortHeadings{An Empirical-Bayes Score for Discrete Bayesian Networks}{Marco Scutari}

\newcommand{\mref}[1]{(\ref{#1})}
\newcommand{\G}{\mathcal{G}}
\newcommand{\GR}{\mathcal{G}_{\mathrm{REF}}}
\newcommand{\D}{\mathcal{D}}
\newcommand{\B}{\mathcal{B}}
\newcommand{\X}{\mathbf{X}}
\newcommand{\XP}{X_i \given \Pi_{X_i}}

\newcommand{\given}{\operatorname{|}}
\newcommand{\Prob}{\operatorname{P}}
\newcommand{\TT}{\Theta_{X_i} \given \Pi_{X_i}}

\begin{document}

\title{An Empirical-Bayes Score for Discrete Bayesian Networks}

\author{\name Marco Scutari \email scutari@stats.ox.ac.uk \\
       \addr Department of Statistics \\
       University of Oxford \\ 
       Oxford, United Kingdom}

\editor{\ldots}

\maketitle

\begin{abstract}%
  Bayesian network structure learning is often performed in a Bayesian setting,
  by evaluating candidate structures using their posterior probabilities for a
  given data set. Score-based algorithms then use those posterior probabilities
  as an objective function and return the \emph{maximum a posteriori} network as
  the learned model. For discrete Bayesian networks, the canonical choice for a
  posterior score is the Bayesian Dirichlet equivalent uniform (BDeu) marginal
  likelihood with a uniform (U) graph prior \citep{heckerman}. Its favourable
  theoretical properties descend from assuming a uniform prior both on the space
  of the network structures and on the space of the parameters of the network.
  In this paper, we revisit the limitations of these assumptions; and we
  introduce an alternative set of assumptions and the resulting score: the
  Bayesian Dirichlet sparse (BDs) empirical Bayes marginal likelihood with a
  marginal uniform (MU) graph prior. We evaluate its performance in an extensive
  simulation study, showing that MU+BDs is more accurate than U+BDeu both in
  learning the structure of the network and in predicting new observations, 
  while not being computationally more complex to estimate.
\end{abstract}


\section{Introduction}

Bayesian networks \citep[BNs;][]{pearl,koller} are a class of statistical models
composed by a set of random variables $\X = \{X_1, \ldots, X_N\}$ and by a
directed acyclic graph (DAG) $\G = (\mathbf{V}, A)$ in which each node in
$\mathbf{V}$ is associated with one of the random variables in $\X$ (they are
usually referred to interchangeably). The arcs in $A$ express direct dependence
relationships among the variables in $\X$; graphical separation of two nodes
implies the conditional independence of the corresponding random variables. In
principle, there are many possible choices for the joint distribution of $\X$;
literature has focused mostly on discrete BNs \citep{heckerman}, in which both
$\X$ and the $X_i$ are multinomial random variables and the parameters of
interest are the conditional probabilities associated with each variable,
usually represented as conditional probability tables. Other possibilities
include Gaussian BNs \citep{heckerman3} and conditional linear Gaussian BNs
\citep{lauritzen}.

The task of learning a BN from data is performed in two steps in an inherently
Bayesian setting. Consider a data set $\D$ and a BN $\B = (\G, \X)$. If we 
denote the parameters of the joint distribution of $\X$ with $\Theta$, we can
assume without loss of generality that $\Theta$ uniquely identifies $\X$ in the
family of distributions chosen to model $\D$ and write 
\begin{align}
\label{eq:lproc}
  &\underbrace{\Prob(\B \given \D) = 
  \Prob(\G, \Theta \given \D)}_{\text{learning}}& &=&
    &\underbrace{\Prob(\G \given \D)}_{\text{structure learning}}& &\cdot&
    &\underbrace{\Prob(\Theta \given \G, \D)}_{\text{parameter learning}}.
\end{align}

\emph{Structure learning} consists in finding the DAG $\G$ that encodes the
dependence structure of the data. Three general approaches to learn $\G$ from
$\D$ have been explored in the literature: constraint-based, score-based and
hybrid. Constraint-based algorithms use conditional independence tests such as
mutual information \citep{itheory} to assess the presence or absence of 
individual arcs in $\G$. Score-based algorithms are typically heuristic search
algorithms and use a goodness-of-fit score such as BIC \citep{schwarz} or the
Bayesian Dirichlet equivalent uniform (BDeu) marginal likelihood 
\citep{heckerman} to find an optimal $\G$. For the latter a uniform (U) prior
over the space of DAGs is assumed for simplicity. Hybrid algorithms combine the
previous two approaches, using conditional independence tests to restrict the
search space in which to perform a heuristic search for an optimal $\G$. For
some examples, see \citet{hiton1}, \citet{larranaga}, \citet{cussens} and
\citet{mmhc}. 

\emph{Parameter learning} involves the estimation of the parameters $\Theta$
given the DAG $\G$ learned in the first step.  Thanks to the Markov property
\citep{pearl}, this step is computationally efficient because if the data are
complete the \emph{global distribution} of $\X$ decomposes into
\begin{equation}
\label{eq:parents}
  \Prob(\X \given \G) = \prod_{i=1}^N \Prob(\XP)
\end{equation}
and the \emph{local distribution} associated with each node $X_i$ depends only
on the configurations of the values of its parents $\Pi_{X_i}$. Note that this
decomposition does not uniquely identify a BN; different DAGs can encode the
same global distribution, thus grouping BNs into equivalence classes
\citep{chickering} characterised by the skeleton of $\G$ (its underlying 
undirected graph) and its v-structures (patterns of arcs of the type 
$X_j \rightarrow X_i \leftarrow X_k$). 

In the remainder of this paper we will focus on discrete BN structure learning
in a Bayesian framework. In Section \ref{sec:bdeu} we will describe the
canonical marginal likelihood used to identify \emph{maximum a posteriori} (MAP)
DAGs in score-based algorithms, BDeu, and the uniform prior U over the space of
the DAGs. We will review and discuss their underlying assumptions and
fundamental properties. In Section \ref{sec:bds} we will address some of their
limitations by introducing a new set of assumptions and the corresponding modified
posterior score, which we will call the \emph{Bayesian Dirichlet sparse} (BDs)
marginal likelihood with a \emph{marginal uniform} (MU) prior. Based on the 
results of an extensive  simulation study, in Section \ref{sec:sim} we will 
show that MU+BDs is preferable to U+BDeu because it is more accurate in 
learning $\G$ from the data; and because the resulting BNs provide better
predictive power than those learned using U+BDeu.

\section{The Bayesian Dirichlet Equivalent Uniform Score (BDeu) with a Uniform
  Prior (U)}
\label{sec:bdeu}

Starting from \mref{eq:lproc}, we can decompose $\Prob(\G \given \D)$ into
\begin{equation}
  \Prob(\G \given \D) \propto \Prob(\G)\Prob(\D \given \G) =
    \Prob(\G)\int \Prob(\D \given \G, \Theta) \Prob(\Theta \given \G) d\Theta
\end{equation}
where $\Prob(\G)$ is the prior distribution over the space of the DAGs and 
$\Prob(\D \given \G)$ is the marginal likelihood of the data given $\G$ averaged
over all possible parameter sets $\Theta$. Using \mref{eq:parents} we can then
decompose $\Prob(\D \given \G)$ into one component for each node as follows:
\begin{equation}
\label{eq:structlearn}
  \Prob(\D \given \G) = \prod_{i=1}^N \Prob(\XP) =
    \prod_{i=1}^N \left[ \int \Prob(\XP, \Theta_{X_i})
    \Prob(\TT) d\Theta_{X_i} \right].
\end{equation}
In the case of discrete BNs, we assume  $\XP \sim \mathit{Multinomial}(\TT)$
where the $\TT$ are the conditional probabilities $\pi_{ijk} = 
\Prob(X_i = k \given \Pi_{X_i} = j)$. We then assume a conjugate prior 
$\TT \sim \mathit{Dirichlet}(\alpha_{ijk})$, $\sum_{jk} \alpha_{ijk} = \alpha_i > 0$
to obtain the posterior $\mathit{Dirichlet}(\alpha_{ijk} + n_{ijk})$ which we
use to estimate the $\pi_{ijk}$ from the counts $n_{ijk}$ observed in $\D$.
$\alpha_i$ is known as the \emph{imaginary} or \emph{equivalent sample size} and
determines how much weight is assigned to the prior in terms of the size of an
imaginary sample supporting it.

Further assuming \emph{positivity} ($\pi_{ijk} > 0$), \emph{parameter
independence} ($\pi_{ijk}$ for different parent configurations are independent),
\emph{parameter modularity} ($\pi_{ijk}$ associated with different nodes are
independent) and \emph{complete data}, \citet{heckerman} derived
a closed form expression for \mref{eq:structlearn}, known as the \emph{Bayesian
Dirichlet} (BD) score:
\begin{equation}
\label{eq:bd}
  \mathrm{BD}(\G, \D; \boldsymbol{\alpha}) = 
  \prod_{i=1}^N \mathrm{BD}(X_i, \Pi_{X_i}; \alpha_i) = 
  \prod_{i=1}^N \prod_{j = 1}^{q_i}
    \left[
      \frac{\Gamma(\alpha_{ij})}{\Gamma(\alpha_{ij} + n_{ij})}
      \prod_{k=1}^{r_i} \frac{\Gamma(\alpha_{ijk} + n_{ijk})}{\Gamma(\alpha_{ijk})}
    \right]
\end{equation}
where $r_i$ is the number of states of $X_i$; $q_i$ is the number of 
configurations of $\Pi_{X_i}$; $n_{ij} = \sum_k n_{ijk}$; and 
$\alpha_{ij} = \sum_k \alpha_{ijk}$. For $\alpha_{ijk} = 1, \alpha_i = r_i q_i$
we obtain the K2 score from \citet{k2}; and for $\alpha_{ijk} = \alpha / (r_i q_i),
\alpha_i = \alpha$ we obtain the \emph{Bayesian Dirichlet equivalent uniform}
(BDeu) score from \citet{heckerman}, which is the most common choice used in
score-based algorithms to estimate $\Prob(\G \given \D)$. It can be shown that
BDeu is score equivalent \citep{chickering}, that is, it takes the same value
for DAGs that encode the same probability distribution. The uniform prior over
the parameters associated with each $\XP$ was justified by the lack of prior
knowledge and widely assumed to be non-informative.

However, there is an increasing amount of evidence that such a set of
assumptions leads to a prior that is far from non-informative and that has a
strong impact on the quality of the learned DAGs. \citet{silander} showed via
simulation that the MAP DAGs selected using BDeu are highly sensitive to the
choice of $\alpha$. Even for ``reasonable'' values such as $\alpha \in [1, 20]$,
they obtained DAGs with markedly different number of arcs, and they showed that
large values of $\alpha$ tend to produce DAGs with more arcs. This is 
counter-intuitive because larger $\alpha$ would normally be expected to result
in stronger regularisation and sparser BNs. \citet{jaakkola} similarly showed
that the number of arcs in the MAP network is determined by a complex interaction
between $\alpha$ and $\D$; in the limits $\alpha \to 0$ and $\alpha \to \infty$
it is possible to obtain both very sparse and very dense DAGs. Furthermore,
they argued that BDeu can be rather unstable for ``medium-sized'' data and small
$\alpha$, which is a very common scenario. \citet{alphastar} approached the
problem from a different perspective and derived an analytic approximation for
the ``optimal'' value of $\alpha$ that maximises predictive accuracy, further
suggesting that the interplay between $\alpha$ and $\D$ is controlled by the
skewness of the $\Prob(\XP)$ and by the strength of the dependence relationships
between the nodes. These results have been analytically confirmed more recently
by \citet{ueno,ueno2}.

As far as $\Prob(\G)$ is concerned, the most common choice is the uniform (U)
distribution $\Prob(\G) \propto 1$; the space of the DAGs grows
super-exponentially in $N$ \citep{harary} and that makes it extremely difficult
to specify informative priors \citep{csprior,mukherjee}. In our previous work
\citep{ba12}, we explored the first- and second-order properties of U and we
showed that for each possible pair of nodes $(X_i, X_j)$
\begin{align}
\label{eq:ba}
  &\overrightarrow{p_{ij}} = \overleftarrow{p_{ij}} \approx \frac{1}{4} + \frac{1}{4(N-1)}&
  &\text{and}&
  &\mathring{p_{ij}} \approx \frac{1}{2} - \frac{1}{2(N-1)},
\end{align}
where $\overrightarrow{p_{ij}} = \Prob(\{X_i \rightarrow X_j\} \in A)$,
$\overleftarrow{p_{ij}} = \Prob(\{X_i \leftarrow X_j\} \in A)$ and
$\mathring{p_{ij}} = \Prob(\{X_i \rightarrow X_j, \linebreak 
X_i \leftarrow X_j\} \not\in A)$. This prior distribution is asymptotically
(marginally) uniform over both arc presence and direction: each arc is present
in $\G$ with probability $\sfrac{1}{2}$ and, when present, it appears in each
direction with probability $\sfrac{1}{2}$. We also showed that two arcs are 
correlated if they are incident on a common node and uncorrelated otherwise
through exhaustive enumeration of all possible DAGs for $N \leqslant 7$ and
through simulation for larger $N$. This suggests that false positives and
false negatives can potentially propagate through $\Prob(\G)$ as well as
$\Prob(\D \given \G)$ and lead to further errors in learning $\G$.

\section{The Bayesian Dirichlet Sparse Score (BDs) with a marginal uniform (MU)
    prior}
\label{sec:bds}

It is clear from the literature review in Section \ref{sec:bdeu} that assuming
uniform priors for $\TT$ and $\G$ can have a negative impact on the quality of
the DAGs learned using BDeu. Therefore, we propose an alternative set of
assumptions; we call the resulting score the \emph{Bayesian Dirichlet sparse}
(BDs) marginal likelihood with a \emph{marginal uniform} (MU) prior.

Firstly, we consider the marginal likelihood BDeu. Starting from \mref{eq:bd},
we can write it as
\begin{equation}
\label{eq:bdeu}
  \mathrm{BDeu}(\G, \D; \alpha) = \prod_{i=1}^N \mathrm{BDeu}(X_i, \Pi_{X_i}; \alpha) = 
  \prod_{i=1}^N \prod_{j = 1}^{q_i}
    \left[
      \frac{\Gamma(r_i \alpha_i^*)}{\Gamma(r_i \alpha_i^* + n_{ij})}
      \prod_{k=1}^{r_i} \frac{\Gamma(\alpha_i^* + n_{ijk})}{\Gamma(\alpha_i^*)}
    \right]
\end{equation}
where $\alpha_i^* = \alpha / (r_i q_i)$. If the positivity assumption is violated
or the sample size $n$ is small, there may be configurations of some $\Pi_{X_i}$
that are not observed in $\D$. In such cases $n_{ij} = 0$ and
\begin{equation}
  \mathrm{BDeu}(X_i, \Pi_{X_i}; \alpha) =
    \prod_{j : n_{ij} = 0}
    \left[
    \cancel{
      \frac{\Gamma(r_i \alpha_i^*)}{\Gamma(r_i \alpha_i^*)}
      \prod_{k=1}^{r_i} \frac{\Gamma(\alpha_i^*)}{\Gamma(\alpha_i^*)}
    }
    \right]
    \prod_{j : n_{ij} > 0}
    \left[
      \frac{\Gamma(r_ i \alpha_i^*)}{\Gamma(r_i \alpha_i^* + n_{ij})}
      \prod_{k=1}^{r_i} \frac{\Gamma(\alpha_i^* + n_{ijk})}{\Gamma(\alpha_i^*)}
    \right].
\end{equation}
This implies that the effective imaginary sample size decreases as the number
of unobserved parents configurations increases, since $\sum_{j : n_{ij} > 0}
\sum_k \alpha_i^* \leqslant \sum_{jk} \alpha_i^* = \alpha$. In turn, the
posterior estimates of $\pi_{ijk}$ gradually converge to the corresponding
maximum likelihood estimates thus favouring overfitting and the inclusion of
spurious arcs in $\G$. Furthermore, the comparison between DAGs with very
different number of arcs may be inconsistent because the respective effective
imaginary sample sizes will be different. \citet{jaakkola} and \citet{silander}
observed both these phenomena, indeed linking them to the interplay between
$\alpha$ and $\D$.

To address these two undesirable features of BDeu we replace $\alpha_i^*$ in
\mref{eq:bdeu}
with
\begin{align}
\label{eq:newprior}
  &\tilde{\alpha}_i = \left\{
    \begin{aligned}
      &\alpha / (r_i \tilde{q}_i)& & \text{if $n_{ij} > 0$} \\
      &0                         & & \text{otherwise.}
    \end{aligned}
  \right.& &\text{where}&
  &\tilde{q}_i = \{ \text{number of $\Pi_{X_i}$ such that $n_{ij} > 0$} \}.
\end{align}
Note that \mref{eq:newprior} is still piece-wise uniform, but now
$\sum_{j : n_{ij} > 0} \sum_k \tilde{\alpha}_i = \alpha$ so the effective
imaginary sample size is equal to $\alpha$ even for sparse data. Intuitively,
we are defining a uniform prior just on the conditional distributions we can
estimate from $\D$, thus moving from a fully Bayesian to an empirical Bayes
score. Plugging \mref{eq:newprior} in \mref{eq:bd} we obtain BDs:
\begin{equation}
  \mathrm{BDs}(X_i, \Pi_{X_i}; \alpha) =
    \prod_{j : n_{ij} > 0}
    \left[
      \frac{\Gamma(r_i \tilde{\alpha}_i)}{\Gamma(r_i \tilde{\alpha}_i + n_{ij})}
      \prod_{k=1}^{r_i} \frac{\Gamma(\tilde{\alpha}_i + n_{ijk})}{\Gamma(\tilde{\alpha}_i)}
    \right]
\end{equation}
If the positivity assumption holds, we will eventually observe all parents 
configurations in the data and thus $\mathrm{BDs}(X_i, \Pi_{X_i}; \alpha) \to
\mathrm{BDeu}(X_i, \Pi_{X_i}; \alpha)$ as $n \to \infty$. Note, however, that
BDs is not score equivalent for finite $n$ unless all  $n_{ij} > 0$. A numeric
example is given below, which also highlights how BDs can be computed in the
same time as BDeu.

\begin{example}
  Consider two binary variables $X_1$ and $X_2$ with data $\D$ comprising
  $x_{11} = 0$, $x_{12} = 0$, $x_{21} = 2$, $x_{22} = 5$ where
  $x_{ij} = \#\{ X_1 = i, X_2 = j \}$. If $\alpha = 1$,
  $\G_1 = \{ X_1 \rightarrow X_2 \}$ and  $\G_2 = \{ X_2 \rightarrow X_1 \}$
  \begin{multline*}
  \mathrm{BDs}(\G_1, \D; 1) = \\
  \left[\frac{\Gamma(1)}{\Gamma(1 + 7)}\frac{\Gamma(\sfrac{1}{2} + 0)\Gamma(\sfrac{1}{2} + 7)}{\Gamma(\sfrac{1}{2})\Gamma(\sfrac{1}{2})}\right]
  \left[\frac{\Gamma(1)}{\Gamma(1 + 7)}
        \frac{\Gamma(\sfrac{1}{2} + 2)\Gamma(\sfrac{1}{2} + 5)}{\Gamma(\sfrac{1}{2})\Gamma(\sfrac{1}{2})}\right]
  = 0.0009,
  \end{multline*}
  \begin{multline*}
  \mathrm{BDs}(\G_2, \D; 1) =
    \left[\frac{\Gamma(1)}{\Gamma(1 + 7)}\frac{\Gamma(\sfrac{1}{2} + 2)\Gamma(\sfrac{1}{2} + 5)}{\Gamma(\sfrac{1}{2})\Gamma(\sfrac{1}{2})}\right] \\
    \left[\frac{\Gamma(\sfrac{1}{2})\Gamma(\sfrac{1}{2})}{\Gamma(\sfrac{1}{2} + 2)\Gamma(\sfrac{1}{2} + 5)}
          \frac{\Gamma(\sfrac{1}{4} + 0)\Gamma(\sfrac{1}{4} + 0)\Gamma(\sfrac{1}{4} + 2)\Gamma(\sfrac{1}{4} + 5)}{\Gamma(\sfrac{1}{4})\Gamma(\sfrac{1}{4})\Gamma(\sfrac{1}{4})\Gamma(\sfrac{1}{4})}\right]
    = 0.0006;
  \end{multline*}
  as a term of comparison the empty DAG $\G_0$ has $\mathrm{BDs}(\G_0, \D) = 0.0009$.
\end{example}

In the general case we have
$\mathrm{BDs}(X_i, \Pi_{X_i}; \alpha) = \mathrm{BDeu}(X_i, \Pi_{X_i}; 
\alpha (q_i / \tilde{q}_i))$ which breaks the score equivalence condition in 
\citet{heckerman} because of the uneven imaginary sample size associated with
each node (like the K2 score). We can interpret $\alpha (q_i / \tilde{q}_i)$ as
an adaptive regularisation hyperparameter that penalises $\XP$ that are not 
fully observed in $\D$, which typically correspond to $X_i$ with a large number
of incoming arcs. Since \citet{jaakkola} showed that BDeu favours the inclusion
of spurious arcs for sparse $\XP$, this adaptive regularisation should lead to
sparser DAGs and reduce overfitting, in turn improving predictive accuracy as
well.

Secondly, we propose a modified prior over for $\G$ with the same aims. We start
from the consideration that score-based structure learning algorithms typically
generate new candidate DAGs by a single arc addition, deletion or reversal. So,
for example
\begin{equation}
\label{eq:hcstep}
  \Prob(\G \cup \{X_j \rightarrow X_i\} \given \D) > \Prob(\G \given \D)
  \Rightarrow \text{accept $\G \cup \{X_j \rightarrow X_i\}$ and discard $\G$}.
\end{equation}
When using the U prior we can rewrite \mref{eq:hcstep} as
\begin{equation}
\label{eq:hcstep2}
  \frac{\Prob(\G \cup \{X_j \rightarrow X_i\} \given \D)}{\Prob(\G \given \D)} =
  \cancel{\frac{\Prob(\G \cup \{X_j \rightarrow X_i\})}{\Prob(\G)}}
  \frac{\Prob(\D \given \G\cup \{X_j \rightarrow X_i\})}
       {\Prob(\D \given \G)} > 1. 
\end{equation}
The fact that U always simplifies is equivalent to assigning equal probabilities
to all possible states of an arc (subject to the acyclicity constraint), say
$\overrightarrow{p_{ij}} = \overleftarrow{p_{ij}} = \mathring{p_{ij}} = \sfrac{1}{3}$ 
using the notation in \mref{eq:ba}. In other words, U favours the inclusion of
new arcs in $\G$ (subject to the acyclicity constraint) as
$\overrightarrow{p_{ij}} + \overleftarrow{p_{ij}} = \sfrac{2}{3}$. Since
\citet{ba12} also showed that arcs incident on a common node are correlated and
may favour each other's inclusion, U may then contribute to overfitting $\G$.

Therefore, we introduce the \emph{marginal uniform} (MU) prior, in which we
assume an independent prior for each arc as in \citet{csprior}, with
probabilities
\begin{align}
  &\overrightarrow{p_{ij}} = \overleftarrow{p_{ij}} = \frac{1}{4}&
  &\text{and}& &\mathring{p_{ij}} = \frac{1}{2}& &\text{for all $i \neq j$}
\end{align}
as in \citet{ba12}. These assumptions make MU computationally trivial to
use: the ratio of the prior probabilities is $\sfrac{1}{2}$ for arc addition,
$2$ for arc deletion and $1$ for arc reversal, for all arcs. Furthermore, arc 
inclusion now has the same prior probability as arc exclusion
($\overrightarrow{p_{ij}} + \overleftarrow{p_{ij}} = \mathring{p_{ij}} = \sfrac{1}{2}$)
and arcs incident on a common are no longer correlated, thus limiting
overfitting and preventing the inclusion of spurious arcs to propagate. However,
the marginal distribution for each arc is the same as in \mref{eq:ba} for large
$N$, hence the name ``marginal uniform''.

\section{Simulation Study}
\label{sec:sim}

\begin{table}[b]
\begin{center}
  \begin{tabular}{|l|r|r|r|c|l|r|r|r|}
  \cline{1-4} \cline{6-9} 
  network    & $N$   & $|A|$ & $p$        & & network    & $N$   & $|A|$ & $p$      \\
  \cline{1-4} \cline{6-9} 
  ALARM      & $37$  & $46$  & $509$      & & HAILFINDER & $56$  & $66$  & $2656$   \\
  ANDES      & $223$ & $338$ & $1157$     & & HEPAR 2    & $70$  & $123$ & $1453$   \\
  ASIA       & $8$   & $8$   & $18$       & & INSURANCE  & $27$  & $52$  & $984$    \\
  CHILD      & $20$  & $25$  & $230$      & & PATHFINDER & $135$ & $200$ & $77155$  \\
  DIABETES   & $413$ & $602$ & $429409$   & & PIGS       & $442$ & $592$ & $5618$   \\
  \cline{1-4} \cline{6-9} 
  \end{tabular}
  \caption{Reference BNs from the BN repository \citep{bnrepository} with the
     respective numbers of nodes ($N$), numbers of arcs ($|A|$) and numbers of
     parameters ($p = |\Theta|$).}
  \label{tab:networks}
\end{center}
\end{table}

\begin{table}[p]
\begin{center}
  \scriptsize
  \begin{tabular}{lr|r|rrr|rr|rr|rr}

    \hline
    \multirow{2}{*}{NETWORK} & \multirow{2}{*}{$n/p$} & \multicolumn{1}{|c|}{BIC} & \multicolumn{3}{|c|}{U + BDeu} & \multicolumn{2}{|c|}{U + BDs} & \multicolumn{2}{|c|}{MU + BDeu} & \multicolumn{2}{|c}{MU + BDs} \\
                             &                        &                           & 1 & $\alpha_S$ & 10            & 1 & 10                         & 1 & 10                          & 1 & 10                         \\ 
    \hline
    \multirow{7}{*}{ALARM} 
            &  0.1 &  55.5 &   78.0 &   80.5 &   112.7 &     64.2 &     87.3 & \bf 53.0 &      83.5 &  \bf 53.0 &       65.5 \\
            &  0.2 &  50.8 &   49.2 &   56.1 &    92.8 &     49.5 &     75.2 & \bf 39.6 &      68.3 &  \bf 39.6 &       56.2 \\
            &  0.5 &  40.8 &   35.5 &   41.9 &    72.0 &     34.9 &     61.5 & \bf 31.3 &      53.5 &  \bf 31.3 &       46.1 \\
            &  1.0 &  33.7 &   31.9 &   37.6 &    62.6 &     29.1 &     51.8 & \bf 27.1 &      49.8 &  \bf 27.1 &       42.1 \\
            &  2.0 &  28.1 &   26.3 &   31.9 &    53.1 &     23.1 &     44.5 & \bf 22.9 &      41.0 &  \bf 22.9 &       36.5 \\
            &  5.0 &  22.6 &   24.4 &   30.1 &    41.6 &     20.9 &     35.0 & \bf 20.4 &      31.6 &  \bf 20.4 &       28.9 \\
    \hline
    \multirow{7}{*}{ANDES} 
            &  0.1 & \bf 367.6 &  642.1 &  997.6 &  1071.0 &    786.5 &   1367.8 &    439.9 &     765.9 &     439.9 &      829.9 \\
            &  0.2 & \bf 278.3 &  450.1 &  686.9 &   773.4 &    522.8 &    957.0 &    313.0 &     560.4 &     313.0 &      572.4 \\
            &  0.5 &     197.4 &  264.9 &  445.3 &   576.0 &    278.4 &    590.7 &    197.1 &     409.1 & \bf 197.1 &      386.2 \\
            &  1.0 &     147.3 &  196.3 &  320.7 &   467.1 &    196.3 &    434.3 &    143.3 &     331.4 & \bf 143.3 &      299.4 \\
            &  2.0 &     116.2 &  142.6 &  246.6 &   388.3 &    139.4 &    345.5 &    109.9 &     280.2 & \bf 109.9 &      243.9 \\
            &  5.0 &      78.3 &  103.5 &  172.2 &   289.2 &    100.8 &    253.6 & \bf 78.2 &     206.5 & \bf  78.2 &      176.5 \\
    \hline
    \multirow{7}{*}{ASIA} 
            &  0.1 &   8.3 &   16.9 &   16.9 &    16.9 &      8.3 &      8.3 &  \bf 8.0 &   \bf 8.0 &   \bf 8.0 &    \bf 8.0 \\
            &  0.2 &   8.6 &   14.1 &   14.1 &    14.1 &      8.5 &      8.5 &      8.5 &       8.0 &       8.5 &    \bf 8.0 \\
            &  0.5 &   8.4 &   10.9 &   11.1 &    14.4 &      8.6 &     10.1 &      8.5 &       8.8 &       8.5 &    \bf 8.0 \\
            &  1.0 &   8.3 &    9.7 &    9.8 &    14.1 &      8.5 &     11.2 &  \bf 8.2 &      10.7 &   \bf 8.2 &        9.6 \\
            &  2.0 &   8.1 &    8.2 &    8.3 &    13.2 &      8.6 &     12.2 &  \bf 7.2 &      10.2 &   \bf 7.2 &        9.6 \\
            &  5.0 &   6.0 &    5.9 &    5.9 &    11.5 &  \bf 5.7 &     10.3 &  \bf 5.7 &       9.7 &   \bf 5.7 &        8.1 \\
    \hline
    \multirow{7}{*}{CHILD} 
            &  0.1 & \bf  28.4 &   39.6 &   44.8 &    51.5 &     38.6 &     46.5 &     31.6 &      36.5 &      31.6 &       33.6 \\
            &  0.2 &      25.2 &   26.9 &   33.0 &    36.0 &     29.9 &     38.1 & \bf 24.6 &      27.5 &  \bf 24.6 &       27.8 \\
            &  0.5 &      21.0 &   21.1 &   23.6 &    25.0 &     21.4 &     24.6 & \bf 18.9 &      21.1 &  \bf 18.9 &       20.7 \\
            &  1.0 &      18.5 &   18.1 &   20.0 &    19.9 &     18.1 &     20.0 & \bf 17.7 &      18.0 &  \bf 17.7 &       17.8 \\
            &  2.0 &      16.1 &   17.0 &   15.6 &    15.4 &     17.0 &     15.4 &     15.8 &      13.4 &      15.8 &   \bf 13.4 \\
            &  5.0 &      14.4 &   14.7 &   12.4 &    12.3 &     14.7 &     12.3 &     12.8 &  \bf  9.4 &      12.8 &   \bf  9.4 \\
    \hline
    \multirow{7}{*}{DIABETES}
            &  0.1 &   484.3 &    399.9 &  522.6 &   444.8 &     387.8 &    378.8 &     400.4 &     429.5 &     400.4 & \bf 378.6 \\
            &  0.2 &   549.4 &    381.0 &  533.2 &   435.0 &     377.5 &    383.2 &     381.0 &     385.6 &     381.0 & \bf 377.3 \\
            &  0.5 &   416.8 &    399.6 &  531.2 &   440.0 &     387.9 &    373.9 &     392.2 &     430.0 &     392.2 & \bf 373.9 \\
            &  1.0 &   412.3 &    373.0 &  530.9 &   420.3 &     375.0 &    372.2 & \bf 368.5 &     415.8 & \bf 368.5 &     372.1 \\
            &  2.0 &   384.8 &    380.9 &  551.6 &   435.3 & \bf 365.6 &    395.7 &     375.7 &     432.8 &     375.7 &     395.0 \\
            &  5.0 &   402.1 &    413.6 &  599.0 &   465.0 & \bf 408.0 &    427.0 &     412.6 &     465.8 &     412.6 &     426.7 \\
    \hline
    \multirow{7}{*}{HAILFINDER}
            &  0.1 &  63.1 &   66.4 &   49.6 &    50.4 &     62.0 & \bf 46.1 &     63.0 &      48.0 &      63.0 &       48.1 \\
            &  0.2 &  48.9 &   54.7 &   44.1 &    40.8 &     50.6 & \bf 36.3 &     51.7 &      38.4 &      51.7 &       45.3 \\
            &  0.5 &  31.9 &   40.0 &   46.9 &    35.1 &     34.7 & \bf 29.9 &     36.8 &      32.1 &      36.8 &       38.5 \\
            &  1.0 &  34.5 &   33.8 &   48.4 &    40.5 &     31.1 &     35.3 &     30.7 &      39.2 &  \bf 30.7 &       35.2 \\
            &  2.0 &  36.4 &   42.0 &   38.8 &    38.4 &     36.0 &     33.3 &     39.0 &      37.1 &      39.0 &   \bf 33.1 \\
            &  5.0 &  16.9 &   24.4 &   27.9 &    21.1 &     18.4 &     15.1 &     21.4 &      19.0 &      21.4 &   \bf 15.0 \\
    \hline
    \multirow{7}{*}{HEPAR2}
            &  0.1 & \bf 143.0 &  183.7 &  226.7 &   269.9 &    192.4 &    292.2 &    149.1 &     209.8 &     149.1 &      210.2 \\
            &  0.2 & \bf 126.6 &  153.7 &  183.8 &   220.2 &    157.4 &    231.1 &    134.3 &     175.6 &     134.3 &      171.9 \\
            &  0.5 & \bf 101.5 &  115.1 &  138.6 &   166.6 &    116.8 &    167.3 &    105.3 &     138.2 &     105.3 &      134.2 \\
            &  1.0 & \bf  85.0 &   93.0 &  108.5 &   132.8 &     94.2 &    128.1 &     88.0 &     109.8 &      88.0 &      105.8 \\
            &  2.0 & \bf  73.9 &   76.5 &   89.3 &   106.6 &     77.5 &    102.3 &     75.0 &      89.0 &      75.0 &       87.0 \\
            &  5.0 & \bf  58.6 &   60.1 &   63.0 &    73.0 &     60.5 &     69.5 &     58.7 &      62.2 &      58.6 &       59.5 \\
    \hline
    \multirow{7}{*}{INSURANCE}
            &  0.1 &      49.5 &     50.6 &   57.1 &    67.8 &     53.0 &     63.0 & \bf 48.5 &      59.7 &  \bf 48.5 &       56.9 \\
            &  0.2 &      46.3 &     47.5 &   55.5 &    63.8 &     49.4 &     60.1 & \bf 45.9 &      58.5 &  \bf 45.9 &       53.7 \\
            &  0.5 &      46.9 &     45.9 &   52.5 &    59.0 &     45.9 &     52.2 & \bf 43.6 &      55.5 &  \bf 43.6 &       49.1 \\
            &  1.0 &      49.8 &     42.3 &   48.0 &    53.6 &     43.7 &     50.2 &     42.3 &      51.0 &  \bf 42.2 &       46.3 \\
            &  2.0 &      46.4 &     42.9 &   48.0 &    53.9 &     42.8 &     49.0 &     43.0 &      51.6 &  \bf 42.6 &       46.2 \\
            &  5.0 &      47.1 &     39.5 &   44.3 &    48.8 & \bf 39.1 &     46.2 &     39.5 &      47.2 &  \bf 39.1 &       44.6 \\
    \hline
    \multirow{7}{*}{PATHFINDER}
            &  0.1 & 278.2 &  269.2 &  398.1 &   345.9 &     250.3 &    292.9 & \bf 237.8 &     309.0 & \bf 237.8 &      257.9 \\
            &  0.2 & 261.0 &  256.2 &  382.7 &   336.2 & \bf 221.1 &    251.2 &     234.2 &     304.6 &     234.2 &      246.8 \\
            &  0.5 & 259.6 &  255.0 &  351.6 &   299.4 & \bf 189.2 &    203.2 &     234.2 &     277.4 &     234.2 &      193.7 \\
            &  1.0 & 240.2 &  242.8 &  342.0 &   289.4 & \bf 171.3 &    182.6 &     220.5 &     264.9 &     220.5 &      173.8 \\
            &  2.0 & 225.9 &  232.3 &  333.9 &   277.8 & \bf 156.9 &    169.7 &     218.2 &     253.2 &     218.2 &      177.8 \\
            &  5.0 & 218.5 &  208.1 &  320.5 &   263.4 &     124.7 &    130.2 &     189.8 &     239.2 &     189.8 &  \bf 119.5 \\
    \hline
    \multirow{7}{*}{PIGS}
            &  0.1 & 130.7 &  114.8 &  155.4 &   203.3 &    116.2 &    163.0 & \bf 106.3 &     166.7 & \bf 106.3 &      146.7 \\
            &  0.2 & 118.0 &  137.1 &  142.3 &   165.6 &    136.7 &    127.5 &     127.5 &     143.2 &     127.5 &  \bf 111.5 \\
            &  0.5 & 131.1 &  132.9 &  134.8 &   142.4 &    131.3 &    110.5 &     122.6 &     126.5 &     122.6 &  \bf  95.4 \\
            &  1.0 & 133.8 &  135.2 &  136.2 &   138.9 &    132.5 &    104.8 &     122.0 &     124.5 &     122.0 &  \bf  91.2 \\
            &  2.0 & 138.7 &  142.8 &  143.6 &   144.8 &    137.2 &    109.0 &     128.2 &     128.8 &     128.2 &  \bf  89.0 \\
            &  5.0 & 149.8 &  155.5 &  155.1 &   156.6 &    150.2 &    116.9 &     140.6 &     140.7 &     140.6 &  \bf  99.2 \\
    \hline
  \end{tabular}
  \caption{Average SHD distance from $\GR$ (lower is better, best in bold).}
  \label{tab:shd}
\end{center}
\end{table}

We assessed BDs and MU on a set of $10$ reference BNs (Table \ref{tab:networks})
covering a wide range of $N$ ($8$ to $442$), $p = |\Theta|$ ($18$ to $77$K) and 
number of arcs $|A|$ ($8$ to $602$). For each BN, we generated $20$ training
samples of size $\sfrac{n}{p} = 0.1$, $0.2$, $0.5$, $1.0$, $2.0$, and $5.0$ (to
allow for meaningful comparisons between BNs with such different $N$ and $p$) and
we learned $\G$ using U+BDeu, U+BDs, MU+BDeu and MU+BDs with $\alpha = 1, 5, 10$
on each sample. For U + BDeu we also considered the optimal $\alpha$ from 
\citet{alphastar}, denoted $\alpha_S$. In addition, we considered BIC as a
term of comparison, since $\mathrm{BIC} \to \log\mathrm{BDeu}$ as $n \to \infty$.
We measured the performance of different scoring strategies in terms of the
quality of the learned DAG using the SHD distance \citep{mmhc} from the $\GR$ of
the reference BN; in terms of the number of arcs compared to $|A_{\mathrm{REF}}|$
in $\GR$; and in terms of predictive accuracy, computing the log-likelihood on a
test set of size $10$K as an approximation of the corresponding Kullback-Leibler
distance. For parameter learning, we used Dirichlet posterior estimates and 
$\alpha = 1$ as suggested in \cite{koller}. All simulations were performed using
the hill-climbing implementation in the \emph{bnlearn} R package \citep{jss09},
which provides several options for structure learning, parameter learning and
inference on BNs (including the proposed MU and BDs). Since $\alpha = 5$ produced
performance measures that are always in between those for $\alpha = 1$ and
$\alpha = 10$, we omit its discussion for brevity.

SHD distances are reported in Table \ref{tab:shd}. MU+BDs outperforms U+BDeu for
all BNs and $\sfrac{n}{p}$ and is the best score overall in $41/60$ simulations.
BIC also outperforms U+BDeu in $39/60$ simulations and is the best score overall
in $9/60$. For U+BDeu, $\alpha = 1$ always results in a lower SHD than $\alpha_s$
and $\alpha = 10$, which is in agreement with \citet{ueno}. The improvement in
SHD given by using BDs instead of BDeu and by using MU instead of U appears to
be largely non-additive; MU+BDs in most cases has the same or nearly the same
SHD as the best between U+BDs and MU+BDeu. However, MU+BDeu is tied with MU+BDs
for the best SHD more often than U+BDs ($21/60$ vs $11/60$) which suggests 
improvements in SHD can be attributed more to the use of MU than that of BDs.
The higher SHD for U+BDeu is a consequence of the higher number of arcs present
in the learned DAGs, shown in Table \ref{tab:narcs}. Both MU+BDs and BIC learn
fewer arcs than U+BDeu in $59/60$ simulations for both $\alpha = 1$ and
$\alpha = 10$; U+BDeu learns too many arcs (i.e., the ratio with
$|A_{\mathrm{REF}}|$ is greater than $1$) in $38/60$ simulations even for
$\alpha = 1$, as opposed to $23/60$ (MU+BDs) and $18/60$ (BIC). As we argued in
Section \ref{sec:bds}, replacing U with MU results in DAGs with fewer arcs for
all BNs and $\sfrac{n}{p}$. Replacing BDeu with BDs results in fewer arcs in 
$32/60$ simulations for $\alpha = 1$ and in $59/60$ for $\alpha = 10$, which 
suggests that the overfitting observed for U+BDeu can be attributed to both U
and BDeu.

\begin{table}[p]
\begin{center}
  \scriptsize
  \begin{tabular}{lr|r|rrr|rr|rr|rr}

    \hline
    \multirow{2}{*}{NETWORK} & \multirow{2}{*}{$n/p$} & \multicolumn{1}{|c|}{BIC} & \multicolumn{3}{|c|}{U + BDeu} & \multicolumn{2}{|c|}{U + BDs} & \multicolumn{2}{|c|}{MU + BDeu} & \multicolumn{2}{|c}{MU + BDs} \\
                             &                        &                           & 1 & $\alpha_S$ & 10            & 1 & 10                         & 1 & 10                          & 1 & 10                         \\
    \hline
    \multirow{7}{*}{ALARM}
            & 0.1 &  0.596 &   1.635 &   1.697 &    2.550 &    1.329 &     1.875 & \bf 1.040 &      1.854 &  \bf 1.040 &       1.351 \\
            & 0.2 &  0.662 &   1.272 &   1.448 &    2.278 &    1.321 &     1.874 & \bf 1.049 &      1.730 &  \bf 1.049 &       1.436 \\
            & 0.5 &  0.746 &   1.174 &   1.290 &    1.993 &    1.213 &     1.775 & \bf 1.060 &      1.605 &  \bf 1.060 &       1.436 \\
            & 1.0 &  0.859 &   1.165 &   1.302 &    1.830 &    1.180 &     1.667 & \bf 1.071 &      1.553 &  \bf 1.071 &       1.426 \\
            & 2.0 &  0.972 &   1.117 &   1.236 &    1.664 &    1.098 &     1.528 & \bf 1.064 &      1.445 &  \bf 1.064 &       1.377 \\
            & 5.0 &  1.092 &   1.098 &   1.208 &    1.457 &    1.086 &     1.386 & \bf 1.061 &      1.286 &  \bf 1.061 &       1.252 \\
    \hline
    \multirow{7}{*}{ANDES} 
            & 0.1 & \bf 1.069 &   1.910 &   3.020 &    3.248 &    2.339 &     4.121 &     1.294 &      2.329 &      1.294 &       2.510 \\
            & 0.2 & \bf 1.032 &   1.550 &   2.303 &    2.570 &    1.764 &     3.115 &     1.129 &      1.926 &      1.129 &       1.963 \\
            & 0.5 &     1.018 &   1.224 &   1.794 &    2.195 &    1.258 &     2.236 & \bf 1.011 &      1.694 &  \bf 1.011 &       1.622 \\
            & 1.0 &     1.011 &   1.156 &   1.556 &    1.999 &    1.154 &     1.898 & \bf 0.991 &      1.593 &  \bf 0.991 &       1.496 \\
            & 2.0 &     1.007 &   1.073 &   1.399 &    1.829 &    1.063 &     1.702 & \bf 0.996 &      1.507 &  \bf 0.996 &       1.399 \\
            & 5.0 & \bf 0.999 &   1.056 &   1.275 &    1.642 &    1.046 &     1.541 &     0.970 &      1.394 &      0.970 &       1.309 \\
    \hline
    \multirow{7}{*}{ASIA} 
            & 0.1 & \bf 0.163 &      2.038 &   2.038 &    2.038 & \bf 0.163 & \bf 0.163 &     0.000 &      0.000 &      0.000 &       0.000 \\
            & 0.2 & \bf 0.338 &      1.669 &   1.669 &    1.669 &     0.163 &     0.163 &     0.163 &      0.000 &      0.163 &       0.000 \\
            & 0.5 &     0.381 &      1.306 &   1.337 &    1.744 &     0.412 & \bf 0.706 &     0.281 &      0.662 &      0.281 &       0.150 \\
            & 1.0 &     0.312 & \bf  1.031 &   1.094 &    1.731 &     0.338 &     0.881 &     0.231 &      1.044 &      0.231 &       0.463 \\
            & 2.0 &     0.544 & \bf  1.025 &   1.031 &    1.769 &     0.762 &     1.325 &     0.544 &      1.238 &      0.544 &       0.838 \\
            & 5.0 &     0.688 & \bf  1.012 &   1.012 &    1.781 &     0.863 &     1.406 &     0.700 &      1.356 &      0.700 &       1.019 \\
    \hline
    \multirow{7}{*}{CHILD}
            & 0.1 &  0.442 &   1.150 &     1.470 &    1.802 &    1.114 &     1.564 &     0.788 &      1.098 &      0.788 &   \bf 0.956 \\
            & 0.2 &  0.588 &   0.894 &     1.250 &    1.366 &    1.014 &     1.444 &     0.744 &      0.992 &      0.744 &   \bf 0.998 \\
            & 0.5 &  0.642 &   0.730 &     1.080 &    1.134 &    0.744 &     1.132 &     0.658 &      0.942 &      0.658 &   \bf 0.950 \\
            & 1.0 &  0.730 &   0.774 & \bf 1.006 &    1.020 &    0.772 &     1.016 &     0.736 &      0.912 &      0.736 &       0.920 \\
            & 2.0 &  0.808 &   0.842 & \bf 1.000 &    0.994 &    0.842 &     0.994 &     0.820 &      0.962 &      0.820 &       0.962 \\
            & 5.0 &  0.914 &   0.908 &     1.046 &    1.034 &    0.908 &     1.034 &     0.898 &  \bf 1.012 &      0.898 &   \bf 1.012 \\
    \hline
    \multirow{7}{*}{DIABETES}
            & 0.1 &  \bf 1.023 &   1.107 &   1.419 &    1.252 &    1.122 &     1.158 &     1.107 &      1.229 &      1.107 &       1.157 \\
            & 0.2 &  \bf 1.065 &   1.115 &   1.447 &    1.237 &    1.136 &     1.169 &     1.115 &      1.200 &      1.115 &       1.168 \\
            & 0.5 &  \bf 1.051 &   1.150 &   1.442 &    1.224 &    1.158 &     1.189 &     1.138 &      1.205 &      1.138 &       1.189 \\
            & 1.0 &  \bf 1.048 &   1.156 &   1.499 &    1.236 &    1.164 &     1.193 &     1.149 &      1.228 &      1.149 &       1.193 \\
            & 2.0 &  \bf 1.083 &   1.176 &   1.539 &    1.281 &    1.192 &     1.264 &     1.167 &      1.276 &      1.167 &       1.262 \\
            & 5.0 &  \bf 1.158 &   1.260 &   1.619 &    1.349 &    1.281 &     1.322 &     1.261 &      1.350 &      1.260 &       1.321 \\
    \hline
    \multirow{7}{*}{HAILFINDER}
            & 0.1 &  0.699 &     0.774 &   1.077 & \bf 0.972 &    0.707 &     0.880 &     0.714 &      0.928 &      0.714 &       0.862 \\
            & 0.2 &  0.782 &     0.901 &   1.098 & \bf 0.977 &    0.839 &     0.880 &     0.852 &      0.942 &      0.852 &       0.873 \\
            & 0.5 &  0.843 &     0.933 &   1.117 & \bf 0.995 &    0.854 &     0.886 &     0.886 &      0.970 &      0.886 &       0.892 \\
            & 1.0 &  0.892 &     0.967 &   1.145 &     1.014 &    0.884 &     0.904 &     0.919 &  \bf 0.992 &      0.919 &       0.901 \\
            & 2.0 &  0.898 & \bf 0.989 &   1.189 &     1.049 &    0.898 &     0.942 &     0.943 &      1.027 &      0.943 &       0.936 \\
            & 5.0 &  0.986 &     1.059 &   1.231 &     1.099 &    0.968 &     0.978 & \bf 1.013 &      1.067 &  \bf 1.013 &       0.977 \\
    \hline
    \multirow{7}{*}{HEPAR2}
            & 0.1 &  0.451 &   0.886 &   1.338 &    1.723 &    0.972 &     1.944 &     0.527 &  \bf 1.198 &      0.527 &       1.202 \\
            & 0.2 &  0.433 &   0.739 &   1.121 &    1.472 &    0.786 &     1.576 &     0.491 &      1.063 &      0.491 &   \bf 1.039 \\
            & 0.5 &  0.467 &   0.654 &   0.962 &    1.250 &    0.680 &     1.252 &     0.498 &  \bf 0.967 &      0.498 &       0.922 \\
            & 1.0 &  0.525 &   0.635 &   0.885 &    1.140 &    0.653 &     1.111 &     0.551 &  \bf 0.908 &      0.551 &       0.875 \\
            & 2.0 &  0.588 &   0.660 &   0.885 &    1.069 &    0.668 & \bf 1.041 &     0.598 &      0.890 &      0.598 &       0.873 \\
            & 5.0 &  0.681 &   0.726 &   0.918 &    1.020 &    0.729 & \bf 0.992 &     0.697 &      0.913 &      0.697 &       0.887 \\
    \hline
    \multirow{7}{*}{INSURANCE}
            & 0.1 &  0.405 &   0.626 &   0.829 & \bf 1.042 &    0.663 &     0.937 &     0.549 &      0.870 &      0.549 &       0.779 \\
            & 0.2 &  0.447 &   0.647 &   0.825 & \bf 1.010 &    0.674 &     0.927 &     0.603 &      0.901 &      0.603 &       0.819 \\
            & 0.5 &  0.535 &   0.689 &   0.859 & \bf 1.048 &    0.700 &     0.906 &     0.662 &      0.962 &      0.662 &       0.830 \\
            & 1.0 &  0.638 &   0.760 &   0.906 &     1.054 &    0.776 &     0.941 &     0.746 &  \bf 0.989 &      0.746 &       0.870 \\
            & 2.0 &  0.723 &   0.806 &   0.942 &     1.103 &    0.811 &     1.012 &     0.799 &  \bf 1.058 &      0.799 &       0.941 \\
            & 5.0 &  0.797 &   0.880 &   1.011 &     1.096 &    0.887 &     1.040 &     0.870 &      1.057 &      0.870 &   \bf 0.994 \\
    \hline
    \multirow{7}{*}{PATHFINDER}
            & 0.1 &  0.815 &   1.154 &   1.862 &    1.591 &    1.062 &     1.337 & \bf 0.961 &      1.391 &  \bf 0.961 &       1.112 \\
            & 0.2 &  0.805 &   1.096 &   1.852 &    1.538 &    0.992 &     1.190 & \bf 0.941 &      1.376 &  \bf 0.941 &       1.044 \\
            & 0.5 &  0.871 &   1.096 &   1.846 &    1.438 &    0.985 &     1.102 & \bf 0.963 &      1.320 &  \bf 0.963 &       1.014 \\
            & 1.0 &  0.864 &   1.081 &   1.871 &    1.477 &    0.965 &     1.068 & \bf 0.951 &      1.343 &  \bf 0.951 &       0.999 \\
            & 2.0 &  0.859 &   1.095 &   1.907 &    1.470 &    0.966 &     1.014 & \bf 1.004 &      1.346 &  \bf 1.004 &       0.958 \\
            & 5.0 &  0.864 &   1.071 &   1.945 &    1.467 &    0.919 &     0.974 & \bf 0.985 &      1.347 &  \bf 0.985 &       0.946 \\
    \hline
    \multirow{7}{*}{PIGS}
            & 0.1 &  1.047 &   1.050 &   1.098 &    1.176 &    1.049 &     1.156 & \bf 1.044 &      1.122 &  \bf 1.044 &       1.112 \\
            & 0.2 &  1.059 &   1.063 &   1.071 &    1.112 &    1.062 &     1.091 & \bf 1.052 &      1.082 &  \bf 1.052 &       1.065 \\
            & 0.5 &  1.062 &   1.065 &   1.067 &    1.079 &    1.063 &     1.060 &     1.059 &      1.066 &      1.059 &   \bf 1.048 \\
            & 1.0 &  1.064 &   1.067 &   1.069 &    1.073 &    1.064 &     1.051 &     1.058 &      1.062 &      1.058 &   \bf 1.044 \\
            & 2.0 &  1.073 &   1.075 &   1.076 &    1.079 &    1.069 &     1.074 &     1.062 &      1.066 &      1.062 &   \bf 1.044 \\
            & 5.0 &  1.078 &   1.085 &   1.085 &    1.086 &    1.079 &     1.061 &     1.074 &      1.074 &      1.074 &   \bf 1.052 \\
    \hline
  \end{tabular}
  \caption{Average number of arcs (rescaled by $|A_{\mathrm{REF}}|$; closer to $1$ is better, best in bold).}
  \label{tab:narcs}
\end{center}
\end{table}

The rescaled predictive log-likelihoods in Table \ref{tab:loglik} show that 
U+BDeu never outperforms MU+BDs for $\sfrac{n}{p} < 1.0$ for the same $\alpha$;
for larger $\sfrac{n}{p}$ all scores are tied, and are not reported for brevity.
U+BDeu for $\alpha_s$ is at best tied with the corresponding score for 
$\alpha = 1$ or $\alpha = 10$. The overall best score is MU+BDs for $7/10$
BNs and BIC for the remaining $3/10$.

\begin{table}[t]
\begin{center}
  \scriptsize
  \begin{tabular}{lr|r|rrr|rr|rr|rr}

    \hline
    \multirow{2}{*}{NETWORK} & \multirow{2}{*}{$n/p$} & \multicolumn{1}{|c|}{BIC} & \multicolumn{3}{|c|}{U + BDeu} & \multicolumn{2}{|c|}{U + BDs} & \multicolumn{2}{|c|}{MU + BDeu} & \multicolumn{2}{|c}{MU + BDs} \\
                             &                        &                           & 1 & $\alpha_S$ & 10            & 1 & 10                         & 1 & 10                          & 1 & 10                         \\ 
    \hline
    \multirow{3}{*}{ALARM}
             & 0.1 &    1.54 &     1.67 &     1.68 &      1.85 &      1.67 &       1.80 &   \bf 1.51 &        1.69 &    \bf 1.51 &         1.60 \\
             & 0.2 &    1.33 &     1.32 &     1.34 &      1.44 &      1.35 &       1.43 &   \bf 1.29 &        1.36 &    \bf 1.29 &         1.34 \\
             & 0.5 &    1.21 &     1.17 &     1.17 &      1.21 &      1.17 &       1.20 &   \bf 1.16 &        1.18 &    \bf 1.16 &         1.17 \\
    \hline
    \multirow{3}{*}{ANDES}
             & 0.1 & \bf 11.12 &    13.14 &    17.56 &     18.59 &     14.75 &      24.40 &      11.90 &       15.77 &       11.90 &        17.77 \\
             & 0.2 & \bf 10.00 &    10.56 &    11.53 &     11.96 &     10.88 &      13.30 &      10.16 &       11.13 &       10.16 &        11.47 \\
             & 0.5 & \bf  9.50 &     9.60 &     9.80 &      9.96 &      9.63 &      10.07 &       9.53 &        9.73 &        9.53 &         9.74 \\
    \hline
    \multirow{3}{*}{ASIA}
             & 0.1 &     0.41 &     0.47 &     0.47 &      0.47 &      0.41 &       0.41 &   \bf 0.39 &    \bf 0.39 &    \bf 0.39 &     \bf 0.39 \\
             & 0.2 &     0.37 &     0.39 &     0.39 &      0.39 &  \bf 0.36 &   \bf 0.36 &   \bf 0.36 &    \bf 0.36 &    \bf 0.36 &     \bf 0.36 \\
             & 0.5 &     0.31 &     0.32 &     0.32 &      0.33 &      0.32 &       0.31 &       0.31 &    \bf 0.30 &        0.31 &         0.31 \\
    \hline
    \multirow{3}{*}{CHILD}
             & 0.1 & \bf 1.82 &     2.03 &     2.19 &      2.30 &      2.07 &       2.31 &       1.91 &        2.04 &        1.91 &         2.00 \\
             & 0.2 & \bf 1.58 &     1.66 &     1.77 &      1.82 &      1.71 &       1.88 &       1.62 &        1.68 &        1.62 &         1.69 \\
             & 0.5 & \bf 1.39 &     1.40 &     1.44 &      1.46 &      1.40 &       1.46 &       1.39 &        1.42 &    \bf 1.39 &         1.42 \\
    \hline
    \multirow{3}{*}{DIABETES}
             & 0.1 & 20.54 & 19.40 & \bf 19.26 &     19.27 & 19.34 & \bf 19.26 & 19.40 & \bf 19.26 &  19.40 & \bf 19.26 \\
             & 0.2 & 19.87 & 19.14 &     19.13 &     19.13 & 19.20 &     19.13 & 19.14 & \bf 19.10 &  19.14 &     19.13 \\
             & 0.5 & 19.24 & 19.05 &     19.03 &     19.04 & 19.10 & \bf 19.00 & 19.05 &     19.04 &  19.05 & \bf 19.00 \\
    \hline
    \multirow{3}{*}{HAILFINDER}
             & 0.1 &     5.31 &     5.31 &     5.24 &      5.23 &      5.30 &       5.23 &       5.31 &    \bf 5.22 &        5.31 &     \bf 5.22 \\
             & 0.2 &     5.13 &     5.13 &     5.09 &      5.09 &      5.12 &   \bf 5.08 &       5.13 &    \bf 5.08 &        5.13 &     \bf 5.08 \\
             & 0.5 &     5.01 &     5.01 &     5.00 &      5.01 &      5.01 &   \bf 4.99 &       5.01 &    \bf 4.99 &        5.01 &     \bf 4.99 \\
    \hline
    \multirow{3}{*}{HEPAR2}
             & 0.1 & \bf 3.49 &     3.73 &     3.98 &      4.24 &      3.81 &       4.68 &       3.58 &        3.90 &        3.58 &         4.04 \\
             & 0.2 & \bf 3.37 &     3.45 &     3.54 &      3.63 &      3.47 &       3.74 &       3.40 &        3.51 &        3.40 &         3.53 \\
             & 0.5 & \bf 3.30 &     3.32 &     3.34 &      3.36 &      3.32 &       3.37 &       3.31 &        3.33 &        3.31 &         3.33 \\
    \hline
    \multirow{3}{*}{INSURANCE}
             & 0.1 &     1.61 &     1.59 &     1.60 &      1.64 &      1.59 &       1.66 &   \bf 1.58 &        1.61 &    \bf 1.58 &         1.62 \\
             & 0.2 &     1.52 & \bf 1.46 & \bf 1.46 &      1.47 &  \bf 1.46 &       1.49 &   \bf 1.46 &        1.47 &    \bf 1.46 &         1.47 \\
             & 0.5 &     1.43 &     1.38 & \bf 1.37 &  \bf 1.37 &      1.38 &       1.38 &       1.38 &    \bf 1.37 &        1.38 &     \bf 1.37 \\
    \hline
    \multirow{3}{*}{PATHFINDER}
             & 0.1 &    2.65 &     2.51 & \bf 2.49 &  \bf 2.49 &      2.50 &   \bf 2.49 &       2.51 &    \bf 2.49 &        2.51 &     \bf 2.49 \\
             & 0.2 &    2.54 & \bf 2.43 & \bf 2.43 &  \bf 2.43 &  \bf 2.43 &   \bf 2.43 &   \bf 2.43 &    \bf 2.43 &    \bf 2.43 &     \bf 2.43 \\
             & 0.5 &    2.45 &     2.39 & \bf 2.38 &      2.39 &      2.39 &   \bf 2.38 &       2.39 &        2.39 &        2.39 &     \bf 2.38 \\
    \hline
    \multirow{3}{*}{PIGS}
             & 0.1 &     33.49 &     33.25 &     33.29 &     33.36 & \bf 33.24 &      33.36 &  \bf 33.24 &       33.31 &   \bf 33.24 &        33.31 \\
             & 0.2 &     33.15 & \bf 33.13 &     33.14 &     33.16 & \bf 33.13 &      33.15 &  \bf 33.13 &       33.14 &   \bf 33.13 &        33.14 \\
             & 0.5 &     33.05 &     33.05 & \bf 33.04 &     33.05 & \bf 33.04 &  \bf 33.04 &  \bf 33.04 &   \bf 33.04 &   \bf 33.04 &    \bf 33.04 \\
    \hline
  \end{tabular}
  \caption{Average predictive log-likelihood (rescaled by $-10000$; lower is
    better, best in bold). $\sfrac{n}{p} = 1.0, 2.0, 5.0$ showed the same 
    value for all scores and are omitted for brevity.}
  \label{tab:loglik}
\end{center}
\end{table}

\section{Conclusions and Discussion}

In this paper we proposed a new posterior score for discrete BN structure 
learning. We defined it as the combination of a new prior over the space of
DAGs, the ``marginal uniform'' (MU) prior, and of a new empirical Bayes 
marginal likelihood, which we call ``Bayesian Dirichlet sparse'' (BDs). Both
have been designed to address the inconsistent behaviour of the classic 
uniform (U) prior and of BDeu explored by \citet{silander}, \citet{jaakkola}
and \citet{ueno} among others. In particular, our aim was to prevent the 
inclusion of spurious arcs.

In an extensive simulation study using $10$ reference BNs we find that MU+BDs
outperforms U+BDeu for all combinations of BN and sample sizes, both in the
quality of the learned DAGs and in predictive accuracy. This is achieved 
without increasing the computational complexity of the posterior score, since
MU+BDs can be computed in the same time as U+BDeu. In this respect, the 
posterior score we propose is preferable to similar proposals in the 
literature. For instance, the NIP-BIC score from \citet{ueno2} and the 
NIP-BDe/Expected log-BDe scores from \citet{ueno3} outperform BDeu but at a
significant computational cost. The same is true for the optimal $\alpha$
proposed by \citet{alphastar} for BDeu, whose estimation requires multiple runs
of the structure learning algorithm to converge. The Max-BDe and Min-BDe scores
in \citet{minbde} overcome in part the limitations of BDeu by optimising for
either goodness of fit at the expense of predictive accuracy, or vice versa.
As a further term of comparison, we also included BIC in the simulation; 
while it outperforms U+BDeu in some circumstances and it is computationally
efficient, MU+BDs is better overall in the DAGs it learns and in
predictive accuracy.

% Acknowledgements should go at the end, before appendices and references
% \acks{Do I have any?}

% \appendix
% \section*{Appendix A.}
% \label{app:theorem}

\end{document}